\documentclass[10pt,twocolumn,letterpaper]{article}

\usepackage{cvpr}
\usepackage{times}
\usepackage{epsfig}
\usepackage{graphicx}
\usepackage{amsmath}
\usepackage{amssymb}
\usepackage[]{algorithm2e}

\usepackage{authblk}
\usepackage[english]{babel}
\usepackage{blindtext}
\usepackage{siunitx}
\usepackage{array}
\usepackage{float}

\usepackage{cvpr}
\usepackage{times}
\usepackage{epsfig}
\usepackage{graphicx}
\usepackage{amsmath}
\usepackage{amssymb}
\usepackage[]{algorithm2e}

\usepackage{authblk}
\usepackage[english]{babel}
\usepackage{blindtext}
\usepackage{siunitx}
\usepackage{array}
\usepackage{float}

\usepackage[breaklinks=true,bookmarks=false]{hyperref}

\cvprfinalcopy 

\def\cvprPaperID{****} 

\begin{document}

\title{Pedestrian Collision Avoidance System (PeCAS): a Deep Learning Approach}

\author{Peetak Mitra\\
University of Massachusetts\\
Amherst, MA 01003\\
{\tt\small pmitra@umass.edu}
}

\maketitle

\begin{abstract}
	We propose a new deep learning based framework to identify pedestrians, and caution distracted drivers, in an effort to prevent the loss of life and property. This framework uses two Convolutional  Neural Networks (CNN), one which detects pedestrians and the second which predicts the onset of drowsiness in a driver, is implemented on a Raspberry Pi 3 Model B+, shows great promise. The algorithm for implementing such a low-cost, low-compute model is presented and the results discussed. 
\end{abstract}

\section{Introduction}
Drowsy Driving is a major cause for accidents in the United States\cite{nhsta}, causing a loss of life and property that is avoidable. As per some estimates \cite{nhsta}, about 10-12\% of road accidents are caused by a drowsy driver. As work hours get longer, and people live far away from their offices and therefore have longer commutes, the problem of drowsy driving has been exacerbated in the recent years. Pedestrians are often victims of such accidents involving drowsy drivers. This work therefore is an attempt to create a low-cost and effect deterrent against pedestrian collision with a distracted/drowsy driver.

Deep Learning has gained an unprecedented attention over the last few years, and deservedly so, as it has introduced transformative results across diverse scientific disciplines including image recognition, natural language processing, cognitive science, computer vision and so on \cite{DL}\cite{DL2}. With explosive growth of available data and computational resources, there have been tremendous growth in interest in applying Deep Learning in various challenging computer vision problems. One important Deep Learning network known as Convolutional Neural Network or CNN, has been at the forefront of such efforts \cite{DL}\cite{DL2}.

There is a huge potential to bringing Deep Learning solutions to the general public via use of cheap and readily available hardware such as Raspberry Pi \cite{rpi}. Raspberry Pi is a low-cost, low-power consuming credit-card sized computer that runs its own version of Linux. It is highly customizable and has been used in the past to do object detection tasks using CNNs \cite{rpi}\cite{rpi2}.

Drowsiness is defined as feeling of being sleepy and lethargic. Since the identification of drowsiness is a multi-component evaluation process including yawning, closure of eyes, tilting of head etc., the current study is only proposed to limit the scope to eye diagnostics. In other words, the CNN model would determine if the vehicle operators’ eyes are open or closed. The framework includes using two cameras on the car dashboard – one looking inward maintaining focus on the vehicle operator and which connects to the drowsiness detection network, the other looking outside to detect pedestrian movements. The metholodogy is discussed in further sections.

While this current work to detect pedestrians can easily be appended to identify other objects, for the scope of this paper we would only seek to identify pedestrians and a limited drowsiness phenomena.

\section{Background/Related Work}
In the past decade drowsiness has been investigated by
various researchers. In \cite{pauly} the authors have used the HOG
SVM approach for drowsiness detection and comparisons are
made with that of a human rater. The popularity of HOG SVM
lies with object detection and not with the eye blink detection.
Further, comparisons with a human rater in highly unreliable
as human decisions are prone to more errors than an
automated one. In \cite{li} have proposed
a smart watch and a headband containing sensors to identify
drowsiness. But this is limited by the choice of the driver who
might not like to wear a headband and might wear a different
watch. In \cite{rn} the authors used a supervised learning method
which needed a highly reliable ground truth. The authors in \cite{perclos} \cite{peetak}
used PERCLOS features for eye status detection. All the
human-defined eye features used in their research were
calculated from eyelid movements. This means the features
are a subset of information provided by eyelid movements.
Hence, extracting and utilizing only those artificial features
can lead to loss of some meaningful information. Researches
in \cite{perclos} have used speaking and smile detection as emotion
detection parameters. In the absence of which a possible
drowsiness condition can occur. Use of such a system can only
make the algorithm more complex to identify drowsiness. Du
et al \cite{du} have used deep learning and facial expression
recognition. But, this approach has the shortcoming of
requirement of a huge amount of data to train a neural network
to work with a high level of accuracy.

Deep learning methods have achieved great successes
in pedestrian detection, owing to its ability to learn discriminative features from raw pixels. However, they treat
pedestrian detection as a single binary classification task,
which may confuse positive with hard negative samples in the figure below.
\begin{figure}[H]
	\centering
	\includegraphics[width=0.7\linewidth]{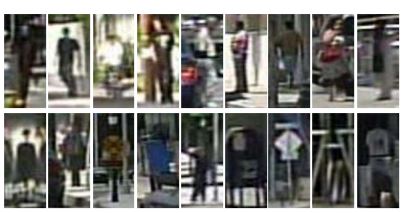}
	\caption{Differences between hard negative and positive samples for pedestrian detector \cite{dalal}}
	\label{fig:1a}
\end{figure}

Pedestrian detection has attracted wide attentions. This problem is challenging
because of large variations and confusions in the human
body and background, as shown in the previous figure, where the
positive and hard negative patches have large ambiguities.
Current methods for pedestrian detection can be generally grouped into two categories, the models based on handcrafted features \cite{Deng} \cite{dalal} and deep models
\cite{dollar}. In the first category, conventional
methods extracted Haar , HOG, or HOG-LBP 
from images to train SVM  or boosting classifiers

The learned weights of the classifier (e.g. SVM) can be
considered as a global template of the entire human body.
To account for more complex poses, the hierarchical deformable part models (DPM)  learned a mixture
of local templates for each body part. Although they are sufficient to certain pose changes, the feature representations
and the classifiers cannot be jointly optimized to improve
performance. In the second category, deep neural networks achieved promising results \cite{dalal}, owing to
their capacity to learn discriminative features from raw
pixels. For example, Dalal et al., learned features
by designing specific hidden layers for the Convolutional
Neural Network (CNN), such that features, deformable
parts, and pedestrian classification can be jointly optimized.

\section{Approach}

 One key aspect to remember is that the implementation of this approach requires a few hardware essentials along with the deep learning model. For the current project, the model was implemented on the Raspberry Pi 3 Model B+ hardware, and two USB webcams (Logitech HD webcam) were used as image acquisition devices to feed in live videos to the Raspberry Pi for detection. 
 
 To implement this framework, there are three essential steps which are listed below.
\begin{itemize}
	\item To train a CNN network to detect pedestrians
	\item To train a CNN network to detect drowsiness
	\item Acquire video feed from two cameras, one looking at the driver (dash cam) and another looking for pedestrians (outward cam). Run the model on both the acquired videos and combine output from the previous two steps in the form of softmax scores, and raise an alarm using the Raspberry Pi hardware when product of their softmax scores crosses a particular threshold.	
\end{itemize}

\begin{figure}[H]
	\centering
	\includegraphics[width=0.7\linewidth]{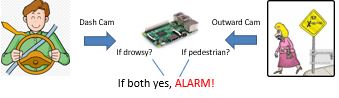}
	\caption{Schematic of Proposed Approach}
	\label{fig:1b}
\end{figure}

While we have this threshold term, which is somewhat arbitrarily set provides another opportunity in a future work to be hyperparameterized. Also another advantage of this approach is that since models are pre-trained, the Raspberry Pi hardware is only making inferences in real time and thereby we can get away with the limited computation of the software. It is to be noted that the entire model is written in MATLAB, because of its superior interfacing with Raspberry Pi hardware. This algorithm could easily be ported using OpenCV into a PyTorch or Tensorflow framework.

\begin{figure}[H]
	\centering
	\includegraphics[width=0.7\linewidth]{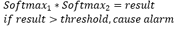}
	\caption{Threshold function that activates the alarm}
	\label{fig:idea}
\end{figure}

\subsection{Data}
\subsubsection{Pedestrian Detector data}

There are many popular pedestrian detector datasets, like Caltech dataset among others. But for this study, we prominently feature data from two sources; UPennFudan People Detector dataset \cite{fudan} as well as INRIA Person dataset \cite{inria}. One of the reasons is because some of the dataset as shown below contains images in a real-world setting with people carrying backpacks and riding bicycles; things that we expect to encounter when we test the model on a college campus like UMass.
The other thing reason is because, the INRIA dataset contains negative examples which are critical to train a robust model some examples of which are shown below. 

\begin{figure} [H]
	\centering
	\includegraphics[width=0.7\linewidth]{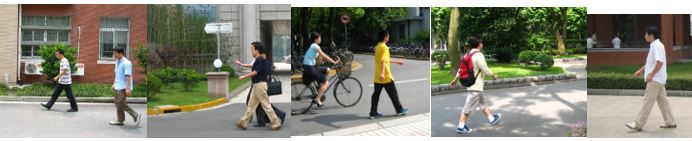}
	\caption{Posittive Examples from Pedestrian detector}
	\label{fig:capture}
\end{figure}

\begin{figure}[H]
	\centering
	\includegraphics[width=0.7\linewidth]{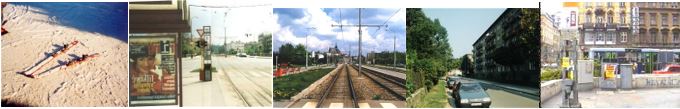}
	\caption{Negative Examples from Pedestrian detector}
	\label{fig:capture2}
\end{figure}

To summarize the above discussion, the data distribution among positive and negative examples look like below. We split the dataset into a train, validation and test dataset based on 0.6, 0.2 and 0.2 distribution.

\begin{tabular}{|c|c|}
	\hline 
	Positiive Examples & Negative Examples \\ 
	\hline 
	12820 & 9000 \\ 
	\hline 
\end{tabular} 

\subsubsection{Drowsiness Detector data}

The dataset for drowsiness detector was harder to obtain because there are not many dataset that deal with the particular problem of eye open/close that I wished to tackle. To generate positive and negative sampling, we obtain data from two different sources. We use the eye gaze dataset publicly available on Kaggle \cite{eyegaze} and closed eyes in the wild dataset \cite{closed} to obtain negative examples. Some examples from the dataset are as below.

\begin{figure}[H]
	\centering
	\includegraphics[width=0.7\linewidth]{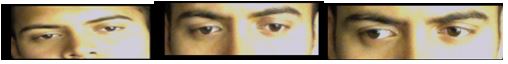}
	\caption{Positive Examples from Eye dataset}
	\label{fig:eyegaze}
\end{figure}

\begin{figure}[H]
	\centering
	\includegraphics[width=0.7\linewidth]{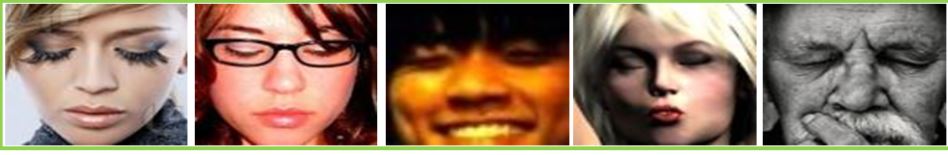}
	\caption{Negative Examples from Eye Dataset}
	\label{fig:close}
\end{figure}

\begin{tabular}{|c|c|}
	\hline 
	Positiive Examples & Negative Examples \\ 
	\hline 
	1950 & 1100 \\ 
	\hline 
\end{tabular}

\subsection{Model}
While there are many approaches one could take including transfer learning and or use of existing pre-trained models for pedestrian detection as well as some relative simple models for SVM, in this current study we have trained the models up from scratch and detail our architecture as well as the results in the following sections.
\subsubsection{Pedestrian Detector model}

For the pedestrian detector model, we started off with a very simple CNN network, consisting of a single convolution layer. But as we kept getting low scores on our validation dataset, we decided to add another layer in the architecture. In the end, the model had two convolutional layers, with stride = 1 and padding = 1.

\begin{figure}[H]
	\centering
	\includegraphics[width=0.2\linewidth]{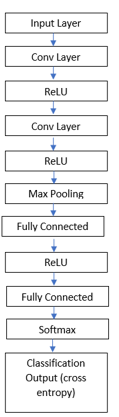}
	\caption{Network architecture for pedestrian model}
	\label{fig:network}
\end{figure}

\subsubsection{Drowsiness Detector model}
For the drowsiness detector model, using a single convolutional layer proved sufficient because the training dataset was simple to decipher. As discussed above, it only contained images of eyes open or eyes closed. In the end, the model had one convolutional layer, with stride = 1 and padding = 1.

\begin{figure}[H]
	\centering
	\includegraphics[width=0.2\linewidth]{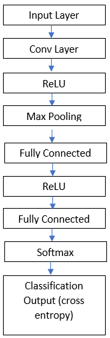}
	\caption{Network architecture for drowsiness model}
	\label{fig:network2}
\end{figure}

\subsubsection{Raspberry Pi 3 additional hardware}

To implement this framework in Raspberry Pi, we used two USB cameras (Logitech Webcam) to acquire the visual feed. The Field of View of the camera was 60 degree with an optical resolution of 1280 x 960 providing a 1.2 Megapixel image quality. The video acquisition was done at 30 frames per second @ 640 x 480. Arguably, a better camera could yield better real time inferences as there would be less noise and more clear features could be detected by the trained network. But for the scope of this project, which serves as a demonstration of this technology, we chose to go ahead with these set of cameras.

\subsection{Algorithm}
While the previous few sections detail the dataset as well as provide a high level approach of the problem, we provide the detailed algorithm in this section. Also, since we were inhibited by the drowsiness dataset (eye open/close), which solely focussed on the eye region we had to perform an additional step before inferences could be made. The step involved using a facial detection algorithm to crop out the eye region from each image obtained from the driver facing camera such that the input image to the drowsiness model mimic as closely to the images it has been trained on. For this step, we used MATLAB's facial recognition algorithm. For future work, this impediment could be overcome by using some other dataset to train the drowsiness model. The training of the models was done on a single GPU computer with NVIDIA GeForce GTX 1080 graphics card.

\textbf{\textit{}}\begin{algorithm}

	\While{video stream is live}{
	 	run pedestrian and drowsiness detector and return softmax scores between 0 and 1 for both detectors. Calculate threshold by a product of the two softmax scores\;
		\eIf{threshold \textgreater 0.2}{
			signal an alarm\;
		}{
			keep making inferences on video inputs\;
		}
	}
	\caption{Algorithm}
\end{algorithm}

\section{Experiment}
The effectiveness of our framework depends on the accuracy of our models as well as the inference times. Since we have trained two different models, we show some of the key results and experiments we conducted. In the figure below, we can see the training and loss chart of the drowsiness detector model. One key factor to note here is that as the model keeps on training, there is a sudden dip in the accuracy shown with corresponding jump in the loss curve which indicates that the local/global minima was overshot.

\begin{figure}[H]
	\centering
	\includegraphics[width=0.7\linewidth]{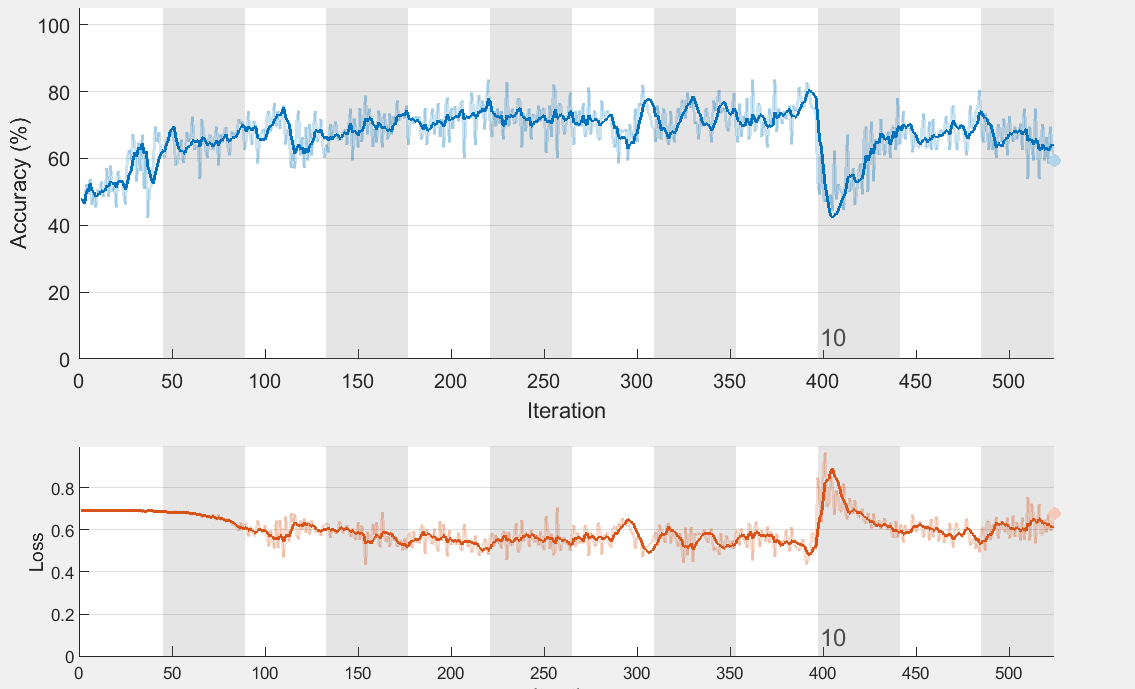}
	\caption{First pass training and loss curve}
	\label{fig:firstpass}
\end{figure}

To overcome this anomaly, we reduced the learning rate which ensure the mimima was found and the accuracy recovered as well as the losses reduced as we kept training the model. 

\begin{figure}[H]
	\centering
	\includegraphics[width=0.7\linewidth]{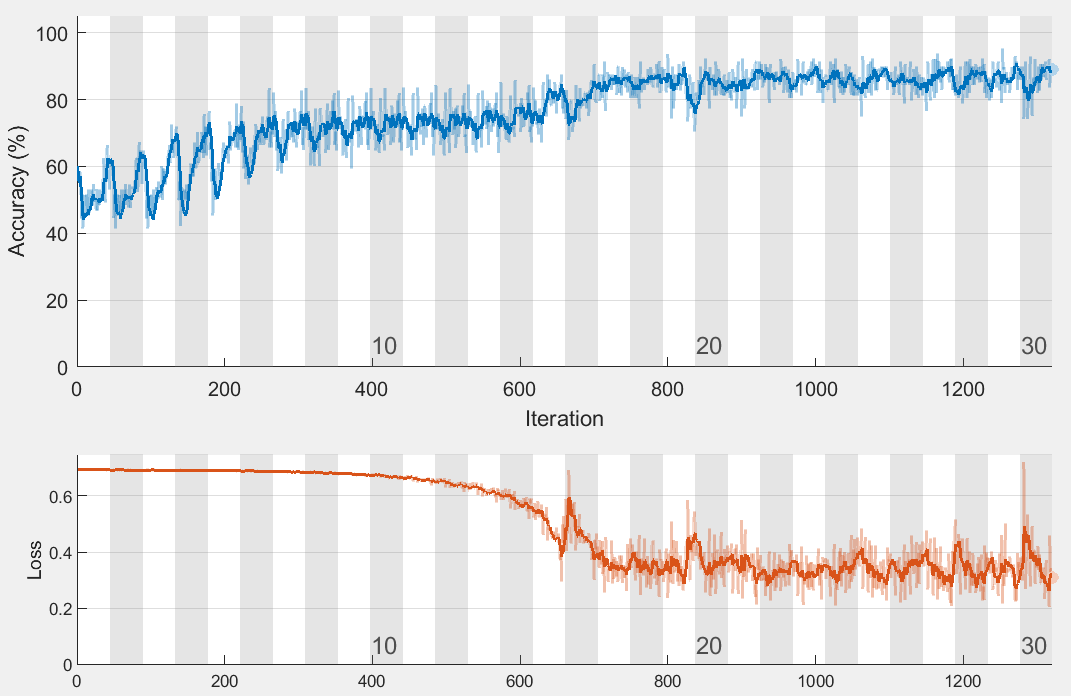}
	\caption{Improved training and loss after lowering learning rate}
	\label{fig:secondpassnetwork}
\end{figure}

In the end, we obtained an accuracy of around 90\% on our test set for the drowsiness detector model which is fairly good and was easy to achieve given our data was singularly focussed on a very small portion of the face.
\begin{figure}[H]
\centering
\includegraphics[width=0.7\linewidth]{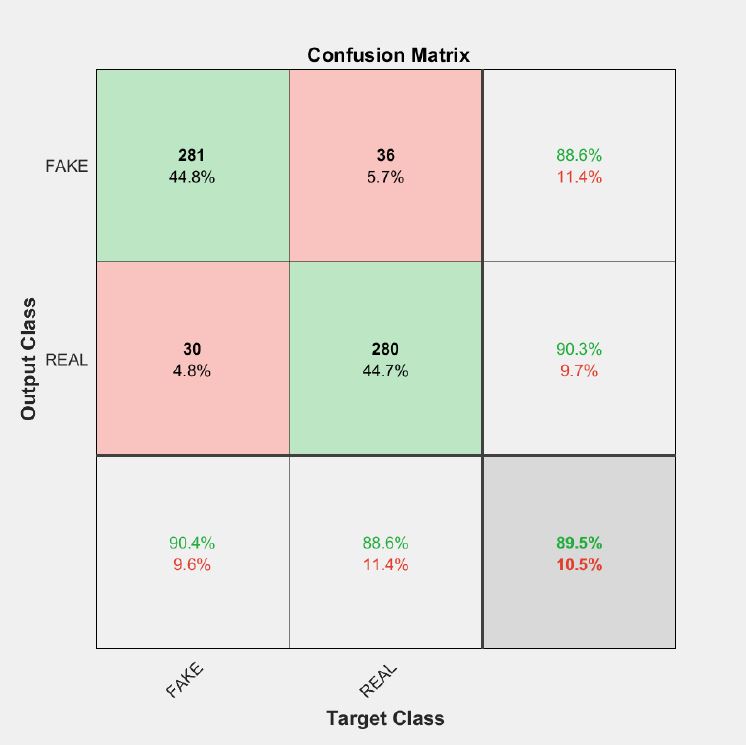}
\caption{Confusion matrix for drowsiness detector}
\label{fig:confusion}
\end{figure}

\begin{figure} [H]
	\centering
	\includegraphics[width=0.7\linewidth]{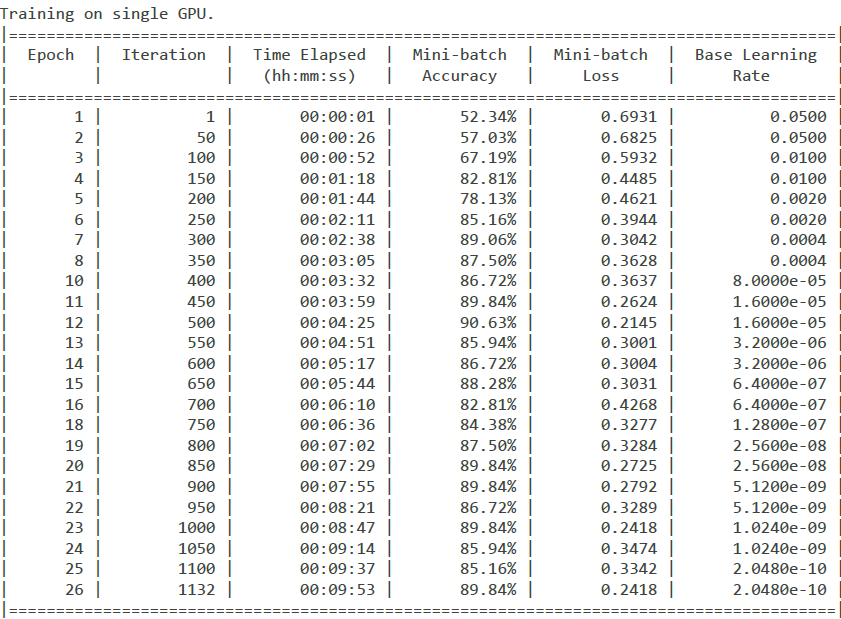}
	\caption{Training details for drowsiness detector}
	\label{fig:untitled}
\end{figure}

For the pedestrian detector, as we discussed earlier we initially began by training a CNN network with one convolutional layer, the results we obtained were not encouraging and hence we added another set of convolutional layer to it which improved the results as we had expected. To better understand the outcome of the model, we plot the precision and recall, which are defined as below.

\[ precision= \frac{TP}{TP + FP}  \]
\[ recall= \frac{TP}{TP + FN}  \]

where TP indicates \textit{true positives}, FP indicates \textit{false positives} and FN indicates \textit{false negatives}. In a nutshell, high precision means that an algorithm returned substantially more relevant results than irrelevant ones, while high recall means that an algorithm returned most of the relevant results.

\begin{figure}[H]
	\centering
	\includegraphics[width=0.7\linewidth]{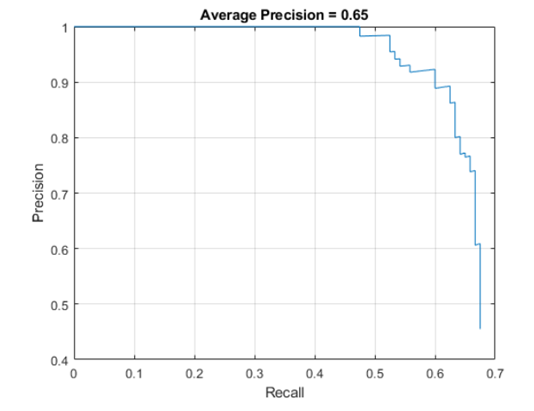}
	\caption{Precision-Recall plot for Pedestrian detector}
	\label{fig:precison}
\end{figure}

We find that the average precision is about 0.65, something that could have been improved further by hyperparameterization or changing the network architecture, which we discuss in the next section.

While the ultimate goal of a model is to look at the correctly predicted classes, it is often helpful to look at some misclassification examples to understand what might be improved in the model. Some examples are shown as below.

In the image below, though clearly has a pedestrian present in the frame the model does not detect one and thereby does not create a bounding box or return any softmax score, which is critical for the second step of the framework. One important aspect to note here would be that the pedestrian is wearing a similar colored clothing as the road and that may have put off the model thereby causing a misclassification.

\begin{figure}[H]
	\centering
	\includegraphics[width=0.7\linewidth]{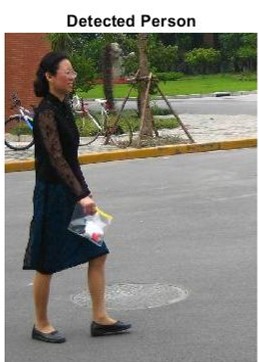}
	\caption{Misclassified example from Pedestrian Detector. The pixels of the clothes match the pixels of the road}
	\label{fig:miss1}
\end{figure}

For the drowsiness detector, since it was trained on an dataset where none of the subjects wore any kind of eyewear like sunglasses or reading glasses, the model is not able to detect the set of eyes in this image and thereby causes misclassification. Another aspect, we found lacking in the model was that it did not do well during low-light conditions which is typical of an interior of a car. This can again be attributed to the images we had in the dataset.
\begin{figure}[H]
	\centering
	\includegraphics[width=0.7\linewidth]{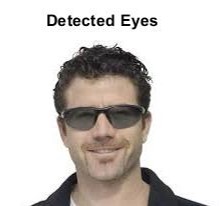}
	\caption{Misclassified example from Eye Detector. Wearing eyewear causes the model to misclassify}
	\label{fig:miss2}
\end{figure}

For the final image, we see that the detector does not detect the person on the bicycle although it detects the nearby pedestrian. This is a major drawback for the pedestrian detector model and can be improved further by tuning some of the hyperparameters or maybe training a deeper network.

\begin{figure}[H]
	\centering
	\includegraphics[width=0.7\linewidth]{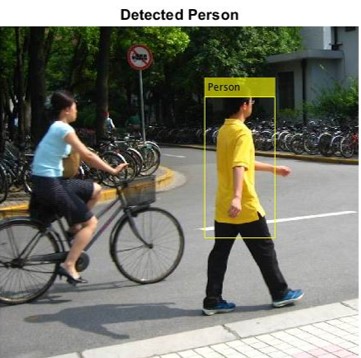}
	\caption{Misclassification of pedestrian detector. Cannot classify people on a bike.}
	\label{fig:miss3}
\end{figure}

Since Raspberry Pi has a limited memory and compute power, it was found that it was typically hard to handle two different live video streams. One of the reasons we chose not to perform hyperparameter tuning was because each pass to the network took a long time and therefore to perform something like a random grid or bayesian optimization would have taken a huge amount of time.

\section{Conclusion}
To conclude, we trained two CNN based networks to detect pedestrians and drowsiness. We demonstrate the applicability of this low-cost, low-compute framework that can be easily applied using Raspberry Pi hardware. The drowsiness detector model got an accuracy of about 90\% and did well in most predictions, but it fell short when the video feed contained an individual wearing any sort of eye wear including transparent reading glasses. One way to overcome such a shortcoming would be to retrain the model or use transfer learning, on another dataset that has images/videos of drivers wearing eye gear. During low light conditions, especially during night time driving, the model based on Raspberry Pi was not able to detect the face and or the eyes. Once again, this can be attributed to the dataset we used to train the network and can be improved further in a future work.

For the pedestrian detector, although we have a decent accuracy with the model we can improve it further by doing a hyperparamter optimization and training it with more data.

The key challenge in implementing this approach is the limited bandwidth of the performance of the Raspberry Pi hardware. To handle two live video feed proved quite a handful for the device and there was a prominent lag in the system that could not be overcome. One way to improve that would include maybe decompressing the video feed to save memory and or taking staggered images. For example, taking a pedestrian image at \textit{time t} and then taking a driver image at \textit{time t+ dt}, and again repeating the cycle. This way there would be no concurrent video feed analysis for the hardware and might ease the job on its limited capabilities.


\begin{thebibliography}{9}
	
	\bibitem{nhsta}
	NHTSA, Report on Drowsiness Driving, accessed 12/4/2018
	\bibitem{DL}
	LeCun, Yann, Yoshua Bengio, and Geoffrey Hinton. "Deep learning." nature 521, no. 7553 (2015): 436.
	\bibitem{DL2}
	Schmidhuber, Jürgen. "Deep learning in neural networks: An overview." Neural networks 61 (2015): 85-117.
	\bibitem{rpi}
	https://www.raspberrypi.org/, accessed 12/19/2018
	\bibitem{rpi2}
 	 https://medium.com/nanonets/how-to-easily-detect-objects-with-deep-learning-on-raspberrypi-225f29635c74, accessed 12/19/2018
 	\bibitem{pauly}
 	Pauly, Leo, and Deepa Sankar. "Detection of drowsiness
 	based on HOG features and SVM classifiers." 2015 IEEE
 	International Conference on Research in Computational
 	Intelligence and Communication Networks (ICRCICN). IEEE,
 	2015.
 	\bibitem{li}
 	Li, G., Lee, B. L.,  Chung, W. Y. (2015). Smartwatch-
 	Based Wearable EEG System for Driver Drowsiness
 	Detection. Sensors Journal, IEEE, 15(12), 7169-7180.
 	\bibitem{rn}
 	R. Nopsuwanchai, Y. Noguchi, M. Ohsuga, Y. Kamakura,
 	and Y. Inoue,“Driver-independent assessment of arousal states
 	from video sequencesbased on the classification of eyeblink
 	patterns,” in Intelligent Transportation.
 	\bibitem{perclos}
 	Rezaee, Khosro, et al. "Real-time intelligent alarm system
 	of driver fatigue based on video sequences." Robotics and
 	Mechatronics (ICRoM), 2013 First RSI/ISM International
 	Conference on. IEEE, 2013
	\bibitem{peetak} 
	Noyes, Matthew A., Peetak Mitra, and Antariksh Dicholkar. "Propagation of Surface-to-Low Earth Orbit Vortex Rings for Orbital Debris Management." In Safety is Not an Option, Proceedings of the 6th IAASS Conference, vol. 715. 2013.
	\bibitem{du}
	Du, Yong, et al. "Driver fatigue detection based on eye
	state analysis." Proceedings of the 11th Joint Conference on
	Information Sciences. 2008.
	\bibitem{dalal}
	N. Dalal and B. Triggs. Histograms of oriented gradients for
	human detection. In CVPR. 2005.
	\bibitem{Deng}
	Y. Deng, P. Luo, C. C. Loy, and X. Tang. Pedestrian attribute
	recognition at far distance. In ACM Multimedia, 2014.
	\bibitem{dollar}
	 P. Dollar, R. Appel, S. Belongie, and P. Perona. Fast feature	pyramids for object detection. TPAMI, 2014.
	\bibitem{fudan}
	https://www.cis.upenn.edu/~jshi/pedhtml/
	\bibitem{inria}
	http://pascal.inrialpes.fr/data/human/
	\bibitem{eyegaze}
	https://www.kaggle.com/4quant/eye-gaze
	\bibitem{closed}
	 http://parnec.nuaa.edu.cn/xtan/data/ClosedEyeDatabases.html
	

	
	
\end{thebibliography}
\end{document}